\documentclass[letterpaper]{article} 
\usepackage{aaai25}  
\usepackage{times}  
\usepackage{helvet}  
\usepackage{courier}  
\usepackage[hyphens]{url}  
\usepackage{graphicx} 
\usepackage{natbib}  
\usepackage{caption} 
\usepackage{amsmath}

\usepackage[ruled,linesnumbered]{algorithm2e}

\usepackage{graphicx}       
\usepackage{multirow}
\usepackage{subfigure}
\usepackage{booktabs}
\usepackage{multirow}

\usepackage{subcaption}
\usepackage{color}
\usepackage{newfloat}
\usepackage{listings}
\DeclareCaptionStyle{ruled}{labelfont=normalfont,labelsep=colon,strut=off} 
\title{Parametric $\rho$-Norm Scaling Calibration}
\author{
    Siyuan Zhang, Linbo Xie\thanks{Corresponding author.}\\
}
\affiliations{
  School of Internet of Things Engineering\\
  Jiangnan University\\
  siyuanz1996@gmail.com, xie\_linbo@jiangnan.edu.cn
}

\usepackage{bibentry}

\begin{document}

\maketitle

\begin{abstract}
Output uncertainty indicates whether the probabilistic properties reflect objective characteristics of the model output. Unlike most loss functions and metrics in machine learning, uncertainty pertains to individual samples, but validating it on individual samples is unfeasible. When validated collectively, it cannot fully represent individual sample properties, posing a challenge in calibrating model confidence in a limited data set. Hence, it is crucial to consider confidence calibration characteristics. To counter the adverse effects of the gradual amplification of the classifier output amplitude in supervised learning, we introduce a post-processing parametric calibration method,  $\rho$-Norm Scaling, which expands the calibrator expression and mitigates overconfidence due to excessive amplitude while preserving accuracy. Moreover, bin-level objective-based calibrator optimization often results in the loss of significant instance-level information. Therefore, we include probability distribution regularization, which incorporates specific priori information that the instance-level uncertainty distribution after calibration should resemble the distribution before calibration. Experimental results demonstrate the substantial enhancement in the post-processing calibrator for uncertainty calibration with our proposed method. 
\end{abstract}

%

\section{Introduction}

Model confidence calibration involves refining uncertainty estimates of the model outputs, enabling more accurate probability predictions that align closely with the objective characteristics of the output uncertainty. With the progressive expansion of model capacity, the modern models often demonstrate inadequately probability distributions. Specifically, these probability outputs display unwarranted over-confidence in comparison to the objective accuracy \cite{moderncalibration}. Furthermore, researchers have identified that achieving high accuracy in classifiers and calibrating the model confidence in baseline are distinct objectives \cite{distinctobjective}. This scenario emphasizes the pressing necessity to rectify the calibration of model output uncertainties in deep learning.

\begin{figure*}[t]
    \centering
    \includegraphics[width=400pt]{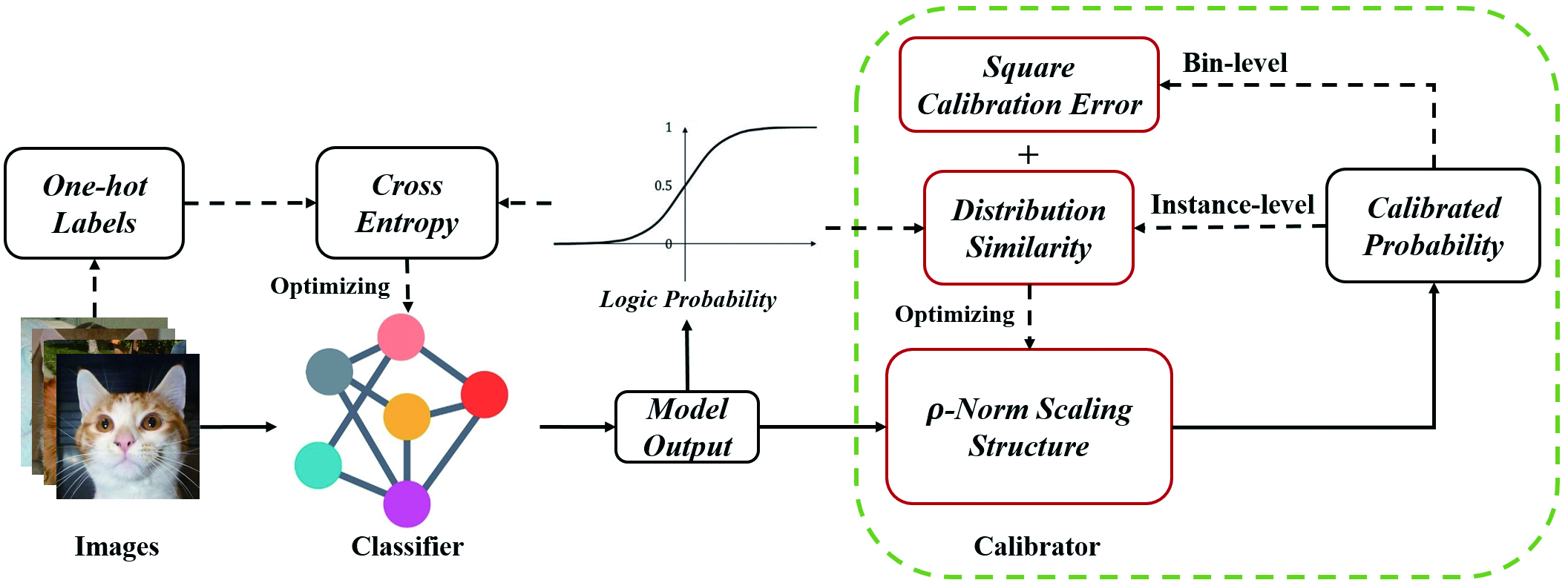}
    \caption{\textbf{Overview of our proposed post-hoc calibrator structure and optimization objective after pipeline of classifier optimization}: (1) Addressing the issue of output magnitude amplification during supervised learning, we introduce a $\rho$-Norm Scaling calibration  within the post-calibration framework. (2) Uncertainty represents the entire dataset statistically, making its optimization prone to losing sample-level information. To address this, we incorporate probabilistic similarity between pre-calibration and post-calibration as a instance-level loss, combined with bin-level loss.}
    \label{papermap}
\end{figure*}

As one of the effective calibration methods, post-calibration methods have recently gained popularity, which operate independently of the model internal optimization, reconstructing and optimizing the output-probability mapping \cite{HB}. In contrast to other calibration techniques, post-processing calibration methods do not necessitate altering the original baseline, thereby preserving the model generalization ability for classification \cite{intraorder}.

Among post-processing calibration methods, parametric techniques like Platt Scaling \cite{platt1999probabilistic}, Temperature Scaling \cite{TSexpressive}, and Beta Calibration \cite{kull2017beta} necessitate parameter optimization using a validation set. The commonly used metric for assessing model confidence, known as Expected Calibration Error (ECE), computing the expected difference between confidence and accuracy within each bin \cite{ECE}. Confidence estimation captures individual sample characteristics, yet it cannot be validated on a per-sample basis. Meanwhile, validating bin-level uncertainty fails to capture the nuances of individual samples. This challenge distinguishes the assessment of model uncertainty and the design and optimization of parametric output-probability mapping. Specifically, the bin-level loss function, such as ECE, is more prone to converge to zero during optimization compared to an instance-level loss function like cross-entropy \cite{AVUO}. This characteristic makes calibrator optimization susceptible to overfitting, impeding generalization. To ensure calibrators learn output-probability mappings effectively, it is essential to incorporate specific priori knowledge to limit the optimization hypothesis space and to design instructive mapping and multi-level optimization objectives.

Inspired by the aforementioned research and questions, we meticulously considered the internal order-preserving property \cite{intraorder} to conserve the inherent uncertainty distribution. Additionally, we assessed the impact of amplitude on the classifier output, supported by prior evidence suggesting that excessive output amplitude might lead to unwarranted calibration error. Building upon this insight, we proposed parametric $\rho$-Norm Scaling calibration, addressing the expressivity limitation in TS \cite{TSexpressive} and mitigating the negative effects by the output amplitudes. Besides, concerning the parameter optimization of the output-probability mapping, we proposed a multi-level loss by introducing a probability distribution similarity regularization into Square Calibration Error. This regularization enhances the correlation between the original probability distribution and the calibrated distribution, aiming to prevent significant deviations from the original probability distribution and retain key properties of the original distribution.

Our main contributions in this work can be summarized as follows: (1) We propose a new family of parametric $\rho$-Norm Scaling calibration model for post-hoc calibration and the corresponding optimization strategy. (2) We provide a new multi-level objective for post-hoc parameter optimization by adding an instance-level regularization between original probability distribution and calibrated probability distribution into bin-level Square Calibration Error. (3) We perform extensive evaluations on multiple datasets and models, and our proposed method achieves state-of-the-art calibration performance.

\section{Related work}
Strategies aimed at calibrating model uncertainty can be classified into the following approaches: Bayesian neural networks, training-based calibration, and post-processing calibration.
The ability of neural network to quantify prediction uncertainty is limited, prompting the consideration of replacing a section of the model structure with a Bayesian inference process \cite{milios2018dirichlet,wen2019batchensemble}. Bayesian neural networks offer several advantages, including ease of implementation, parallelization feasibility, minimized hyperparameter adjustments, and the ability to offer precise estimates of predictive uncertainty \cite{gal2016dropout,calandra2016manifold,ECE,distinctobjective,bayesian3}. Besides, training-based calibration methods have been explored to mitigate miscalibration risks in supervised learning. These approaches may involve techniques such as pre-training \cite{hendrycks2019using}, data augmentation \cite{thulasidasan2019mixup}, label smoothing \cite{menon2020distillation}, weight decay \cite{moderncalibration}, and more. Furthermore, Tao investigated the limitations of early stopping and devised solutions to overfitting within specific network blocks concerning calibration metrics \cite{tao2023calibrating}. The connection between model structure sparsity and model calibration was examined in \cite{lei2022calibrating}. Additionally, researchers have proposed some innovative loss functions, such as MMCE \cite{kumar2018trainable}, Correctness Ranking Loss \cite{moon2020confidence}, CALS \cite{liu2023class}, Focal loss \cite{lin2017focal,Dualfocal,Calibratingfocal}, and FLSD \cite{ghosh2022adafocal}, which simultaneously consider classification accuracy and confidence calibration. However, an excessive focus on model calibration during training might detrimentally affect overall model accuracy improvement. Furthermore, the calibration during training may compromise the efficacy of post-processing calibration methods \cite{compromiseacc}.

Post-processing calibration refers to reconstructing the output-probability mapping. One of its key advantages lies in decoupling classifier accuracy from calibration, thereby maintaining the original generalization without necessitating alterations to its training strategy. In the era preceding the ascendancy of deep learning, non-parametric post-processing calibration methods such as Histogram Binning (HB) \cite{HB}, Isotonic Regression (IR) \cite{IR} and Bayesian processes \cite{gal2016dropout} were prevalent. Unlike non-parametric methods, which calibrate a model confidence distribution using nonlinear logic, parametric calibration methods focus on establishing a parametric structure by learning from finite samples. Some commonly utilized parametric calibration structures include Platt Scaling \cite{platt1999probabilistic}, Temperature Scaling \cite{MultidomainTS,TSexpressive}, Beta Calibration \cite{kull2017beta}. 
Typically, parameters are learned through grid search or gradient-based optimization of Negative Log-Likelihood (NLL) \cite{hastie2009elements}. However, direct optimization of NLL often may compel the model output towards one-hot distribution, deviating from the intended calibration logic. To address this problem, Krishnan introduced the AvUC loss function \cite{AVUO}. Subsequently, Karandikar proposed soft calibration objective for optimizing the calibrator parameters \cite{Softcalibration}. Additionally, in terms of mapping structure, Kull extended Beta Calibration and introduced Dirichlet calibration \cite{kull2019beyond}. Considering the flexibility of calibration mapping, Wang introduced Shape-Restricted Polynomial Regression as a parametric calibration method \cite{wang2019calibrating}. Furthermore, some studies propose a class of accuracy-preserving mappings \cite{TSexpressive, intraorder}.

\section{Methodology}\label{sec3}
\subsection{Problem Formulation}
Considering a dataset  $\left\{ {({x^i},{y^i})} \right\}_{i = 1}^N \subset {{\bf{R}}^n} \times {{\bf{R}}^m}$ and classifier $f$  maps  $x$ to the outputs ${z_j},j = 1, \ldots ,m$  on $m$  classes and $k = \arg {\max _j}{z_j}$. The ground-truth  $y$ and predicted labels $\hat y$  are formulated in one-hot format where  ${y_c} = 1$ and ${\hat y_k} = 1$, where $c$ represents the truth class. The confidence score of the predicted label in baseline is  $\hat p = \max {s_j}(z),j = 1, \ldots ,m$, where $s\left(  \cdot  \right)$  represents Softmax mapping ${R^m} \to {R^m}$. However, Softmax mapping probabilities are not accurately reflected in the properties of model output \cite{moderncalibration}. To address this, the calibrator $g\left(  \cdot  \right)$  is introduced as a new output-probability mapping $g:z \to p$ for probability calibration. \\

\noindent\textbf{Confidence Calibration:} Perfect calibration of neural network can be realized when the confidence score reflects the real probability that the sample is classified correctly. Formally, the perfectly calibrated network satisfied ${\rm P}\left( {\hat y = y|\hat p = p} \right) = p$ for all $p \in \left[ {0,1} \right]$. However, in practical applications, the sample is divided into $M$  bins  $\left\{ {{D_b}} \right\}_{b = 1}^M$. The limited availability of data restricts our ability to accurately estimate the calibration error. According to their confidence scores and the calibration error, an approximation is calculated for each bins  $\left\{ {{D_b}} \right\}_{b = 1}^M$.  ${D_b}$ contains all sample with  $\hat p \in \left[ {\frac{b}{M},\frac{{b + 1}}{M}} \right)$. Average confidence is computed as $conf\left( {{D_b}} \right) = \frac{1}{{\left| {{D_b}} \right|}}\sum\nolimits_{i \in {D_b}} {{{\hat p}^i}} $  and the bin accuracy is computed as  $acc\left( {{D_b}} \right) = \frac{1}{{\left| {{D_b}} \right|}}\sum\nolimits_{i \in D_b} {\rm I} \left( {y_c^i = \hat y_c^i} \right)$. ECE \cite{ECE} is calculated as follows.
\begin{equation}\label{EQ1}   
    ECE = \sum\limits_{b = 1}^M {\frac{{\left| {{D_b}} \right|}}{N}} \left| {acc\left( {{D_b}} \right) - conf\left( {{D_b}} \right)} \right|
\end{equation}

To develop an effective output-probability mapping, we design the calibrator based on the relationship between output amplitude and confidence level. Additionally, we formulate the multi-level objective for calibrators parameter based on the characteristics of calibration error. \\

\begin{figure*}[t]
    \centering
    \subfigure[Amplitude  without weight decay]{\includegraphics[width=125pt]{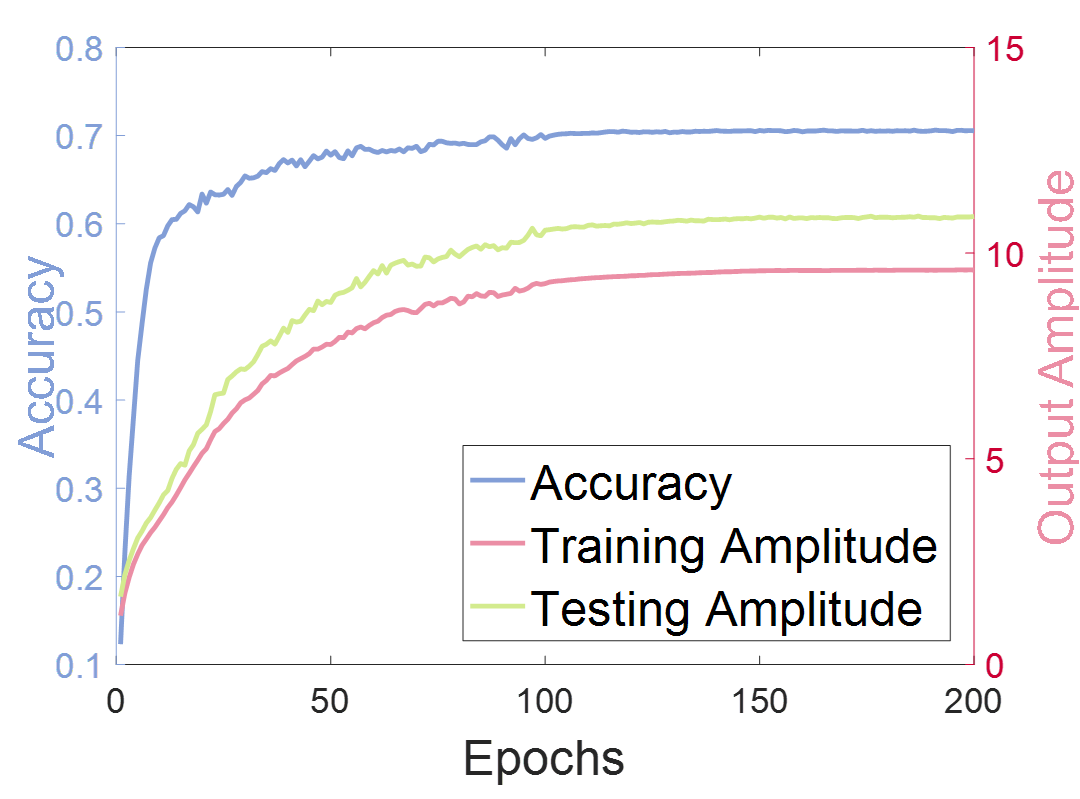}}
    \subfigure[Amplitude with weight decay]{\includegraphics[width=125pt]{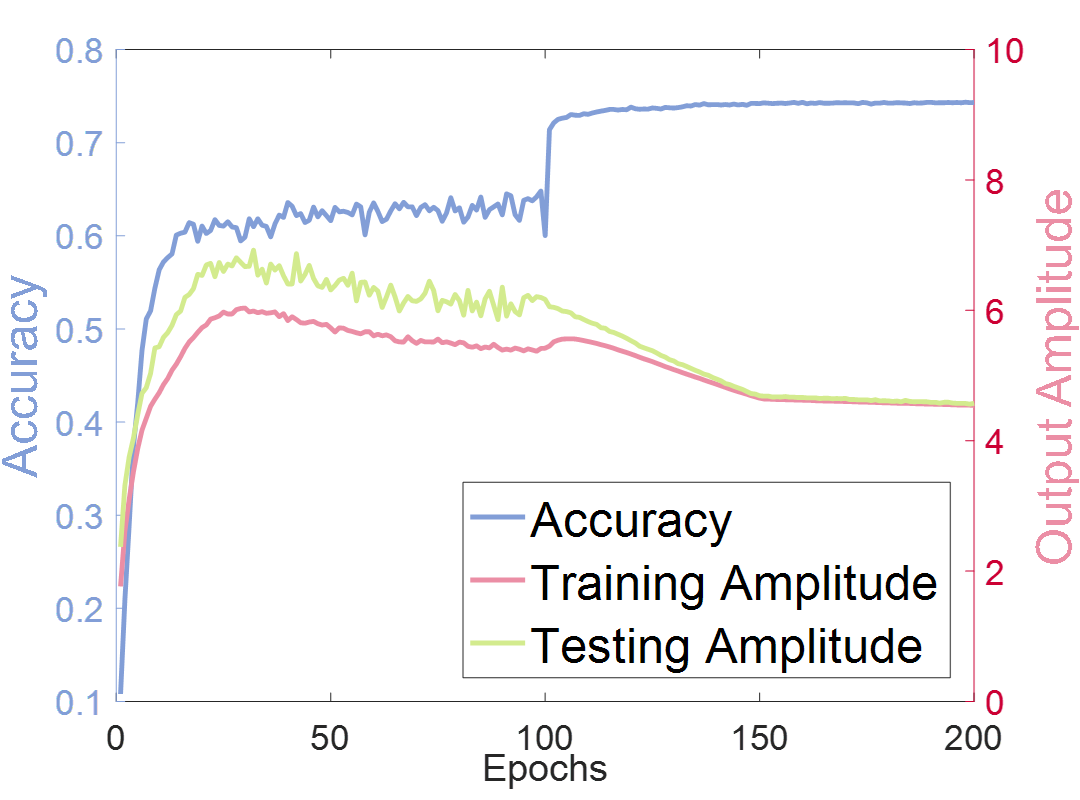}}
    \subfigure[Confidence histogram]{\includegraphics[width=100pt]{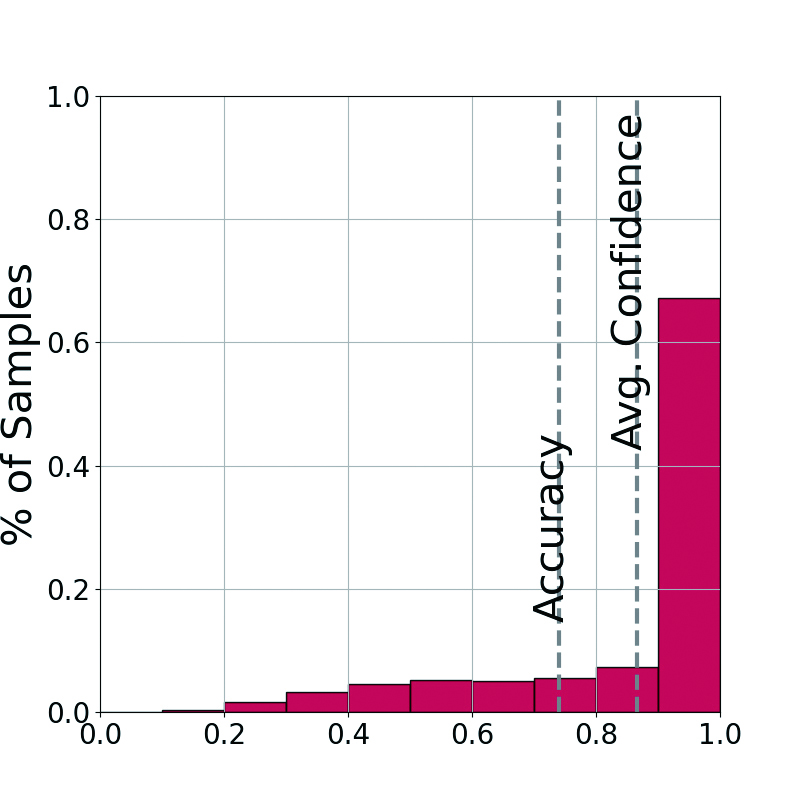}}
    \subfigure[Amplitude histogram]{\includegraphics[width=125pt]{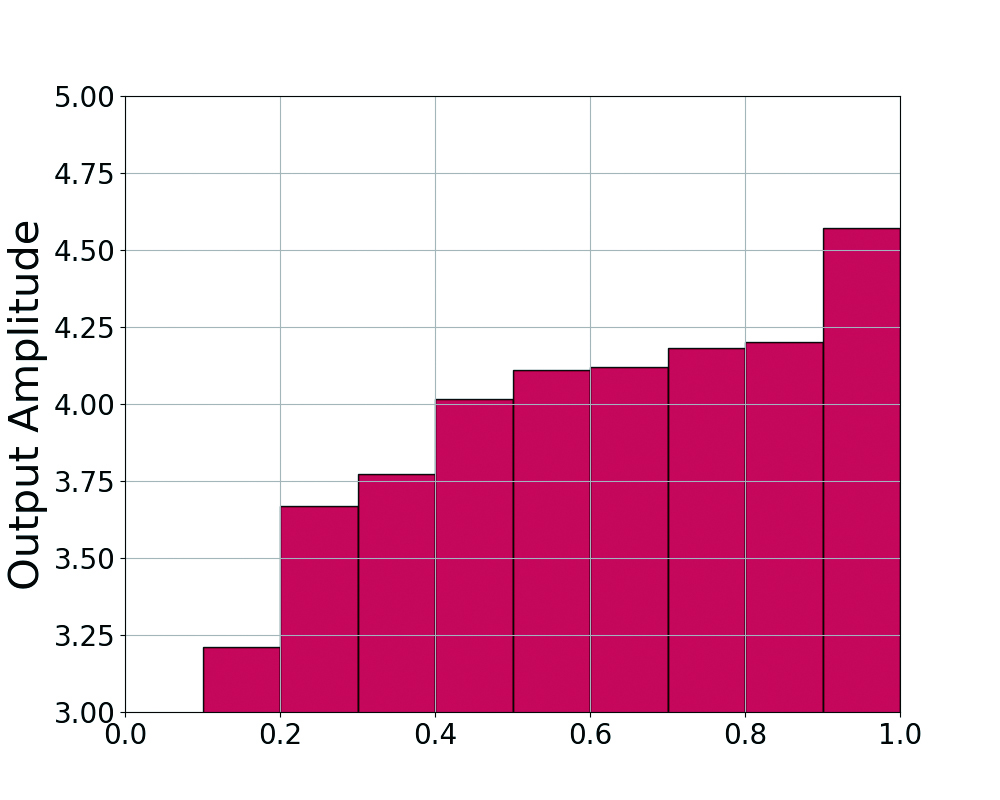}}
    \caption{\textbf{Amplitude changes in classifier optimization.} In these figures, the overall output magnitude of all samples is defined as  $\frac{1}{{Nm}}\sum\nolimits_{i = 1}^N {{{\left\| {{z^i}} \right\|}_2}} $. During the supervised learning of the classifier, the output magnitude follows a specific pattern.  (a) illustrates that in the absence of weight decay, the output amplitude steadily increases throughout the optimization process. Although this trend is alleviated in the presence of weight decay, as depicted in  (b), the final magnitudes  exhibit a positive correlation with the overall confidence distribution, shown in (c) and (d). }
    \label{Norm}
\end{figure*}

\noindent\textbf{Output Magnitude:} Our postulation suggests that overconfidence of modern deep model arises from the utilization of Softmax cross entropy optimization, particularly in high-capacity models, leading to an amplification of the output amplitude, as shown in Fig. \ref{Norm}, where the amplitude of a single sample's output is defined as ${{\left\| {{z^i}} \right\|}_2}$. The unsaturated regions of the original Softmax are present exclusively when the differences between the category outputs are relatively small. Consequently, a significant portion of samples tends to have probabilities that fall within the saturation region of Softmax, resulting in probability outputs close to 1. The similar conclusion can be found in \cite{DecouplingMaxLogit,mitigatinglogit}, where it is suggested that the magnitude of neural network output can be a culprit. \\

\noindent\textbf{Calibration Characteristic:} The optimization of the calibration error within the post-calibration structure appears to be challenging due to hard binning operation, as discussed in literature \cite{Softcalibration}. We believe that another practical obstacle arises from the nature of the uncertainty estimation, which serves as collective binning properties for multiple samples $\sum\nolimits_{i \in {D_b}} {{{\hat p}^i}} = \sum\nolimits_{i \in D_b} {\rm I} \left( {y_c^i = \hat y_c^i} \right)$, rather than providing individual metric for each sample \cite{instanceloss,ecenotgood}. Calibration error as loss metric disregards numerous sample-level output-probability mapping relationships. Consequently, the calibrated distribution significantly diverges from the original distribution, as shown in Fig. \ref{Condidence distribution}. 
\subsection{Parametric $\rho$-Norm Scaling}
To regulate the influence of output amplitude on the scaling calibration and propose accurate calibrator, we enhance the expressive power by incorporating a parameterized  $\rho$-norm normalization term into the output-probability mapping. The adopted calibration structure is represented below:
\begin{equation}\label{EQ3}
    {g_c}\left( z \right) = \frac{{{e^{{r_c}}}}}{{\sum\nolimits_{j = 1}^m {{e^{{r_j}}}} }}
\end{equation}
where  ${r_j}\left( z \right) = \frac{{{z_j}}}{{\gamma {{\left\| z \right\|}_\rho } + \beta }}$,  $\gamma  > 0$ and  $\beta  > 0$.  ${\left\|  \cdot  \right\|_\rho }$ represents  $\rho$-norm, where  ${\left\| z \right\|_\rho } = {\left( {{z_1}^\rho  + {z_2}^\rho  +  \cdots  + {z_m}^\rho } \right)^{{1 \mathord{\left/
 {\vphantom {1 \rho }} \right.
 \kern-\nulldelimiterspace} \rho }}}$. $\rho $  is defined as a learnable parameter in the algorithm that is used to control a learnable norm space to regulate the large output magnitude. In supervised learning, classifiers often produce outputs with substantial magnitudes, particularly for overconfident samples. When these outputs have excessively large magnitudes and are fed into a calibrator, they often fall within the saturation interval where the Softmax output converges to 1. Consequently, calibrating samples with such high-amplitude outputs becomes insensitive.

\newtheorem{proposition}{Proposition}
\begin{proposition} \label{Bound of RMSNorm Softmax}
 For any model output $z$  and the probability by mapping of ${g_c} = \frac{{{e^{{r_c}}}}}{{\sum\nolimits_{j = 1}^m {{e^{{r_j}}}} }}$  where  ${r_j}\left( z \right) = \frac{{{z_j}}}{{\gamma {{\left\| z \right\|}_\rho }}}$, the following inequalities holds.
 \begin{equation}
 \begin{split}
     \frac{1}{{\left( {m - 1} \right){e^{\frac{1}{\gamma }{{\left( {\frac{1}{{{{\left( {m - 1} \right)}^{{1 \mathord{\left/
 {\vphantom {1 {\rho  - 1}}} \right.
 \kern-\nulldelimiterspace} {\rho  - 1}}}}}} + 1} \right)}^{{{\rho  - 1} \mathord{\left/
 {\vphantom {{\rho  - 1} \rho }} \right.
 \kern-\nulldelimiterspace} \rho }}}}} + 1}} \le g \\ \le \frac{1}{{\left( {m - 1} \right){e^{ - \frac{1}{\gamma }{{\left( {\frac{1}{{{{\left( {m - 1} \right)}^{{1 \mathord{\left/
 {\vphantom {1 {\rho  - 1}}} \right.
 \kern-\nulldelimiterspace} {\rho  - 1}}}}}} + 1} \right)}^{{{\rho  - 1} \mathord{\left/
 {\vphantom {{\rho  - 1} \rho }} \right.
 \kern-\nulldelimiterspace} \rho }}}}} + 1}}
 \end{split}
 \end{equation}
\end{proposition}

\begin{proposition} \label{Accuracy preserving}
For any $\gamma  > 0$  and $\beta  > 0$  of  $\rho $-Norm Scaling ${g}\left( z \right)$, the classification accuracy based one-versus-all classification keeps unchanged after the output-to-probability mapping.
\end{proposition}

The results in Proposition \ref{Bound of RMSNorm Softmax} show that  $\rho $-norm is able to restrict the confidence interval to prevent the confidence from a one-hot distribution, yielding a smoother confidence distribution. When  $\gamma  \to  + \infty $, the calibrated probabilities  $g$ tend to ${1 \mathord{\left/
 {\vphantom {1 m}} \right.
 \kern-\nulldelimiterspace} m}$. When  $\gamma  \to  + 0$, all probabilities are adjusted to  $\left[ {0,1} \right]$. Furthermore, Proposition \ref{Accuracy preserving} establishes that the  $\rho $-Norm Scaling structure maintains decision invariance, ensuring that the pre-calibration classification results match the calibrated probability one-versus-all classification outcomes. This property serves as specific priori knowledge, preserving dataset distributional properties during optimization, aligning with calibration logic. Decision invariance is a crucial consideration in calibration structure design \cite{TSexpressive,intraorder}.

\begin{proposition} \label{Accuracy preserving2}
For any output-to-probability mapping  ${g_j}\left( z \right) = \frac{{{e^{{z_j}\sigma \left( z \right)}}}}{{\sum\nolimits_{j = 1}^m {{e^{{z_j}\sigma \left( z \right)}}} }}$, if the function $\sigma \left( z \right) > 0$  holds for any  $z$, accuracy of model based one-versus-all classification decision-making keeps unchanged after the output-to-probability mapping.
\end{proposition}

Proposition \ref{Accuracy preserving2} extends the conclusion in Proposition \ref{Accuracy preserving} by presenting a general mapping-based criterion for satisfying decision invariance, for mapping design. When the function $\sigma \left( z \right)$ is defined as a single hyperparameter, this mapping degenerates into naive Softmax mapping with a temperature coefficient. Apart from calibrators, classifier optimization with different $\sigma \left( z \right)$ deserves further exploration. 

\begin{algorithm}[t]
  \SetAlgoLined
  \KwData{Validation set  $\left\{ {\left. {\left( {{x_i},{y_i}} \right)} \right|i = 1, \ldots ,N} \right\}$; Classifier ${f}$; Learning rate $\lambda $; Batch size  ${N_B}$.}
  \KwResult{$\rho$-Norm Scaling calibrator ${g}\left( z, \rho^*, \theta^* \right)$}

  Initialize $\theta$, $\theta^*$, $\rho^*$, $ECE^*$\;
  $z \gets {f}\left( x \right)$\;
  \While{$\rho  \in \left\{ {1, \ldots ,3} \right\}$}{
        \While{$t < T_{max}$}{
            $D_{N_B} \gets$ $\left\{{ z^i }\right\}_i^{N_B}$ \;
            ${l_{SCE}} \gets$ Computing by Eq.(\ref{eqsce})\;
            ${l_{KL}} \gets$ Computing by Eq.(\ref{eqkl})\;
            ${\theta} \gets {\theta } - \lambda \frac{{\partial {l}}}{{\partial {\theta}}}$\;
}
        $ECE_{val}$ $\gets$ $ECE$ on validation set by Eq.(\ref{EQ1})\;
    \If{$ECE_{val} < ECE^*$}{
         $\rho^* \gets \rho $\;
         $\theta^* \gets \theta $\;
      $ ECE^* \gets ECE_{val} $\;
      }
    }
  \caption{$\rho$-Norm Scaling Post-hoc Calibrator}\label{algorithm1}

\end{algorithm}

\begin{figure*}[h]
    \centering
    \subfigure[Uncalibrated]{\includegraphics[width=80pt]{Res34_CIFAR100_original_1.jpg}\includegraphics[width=80pt]{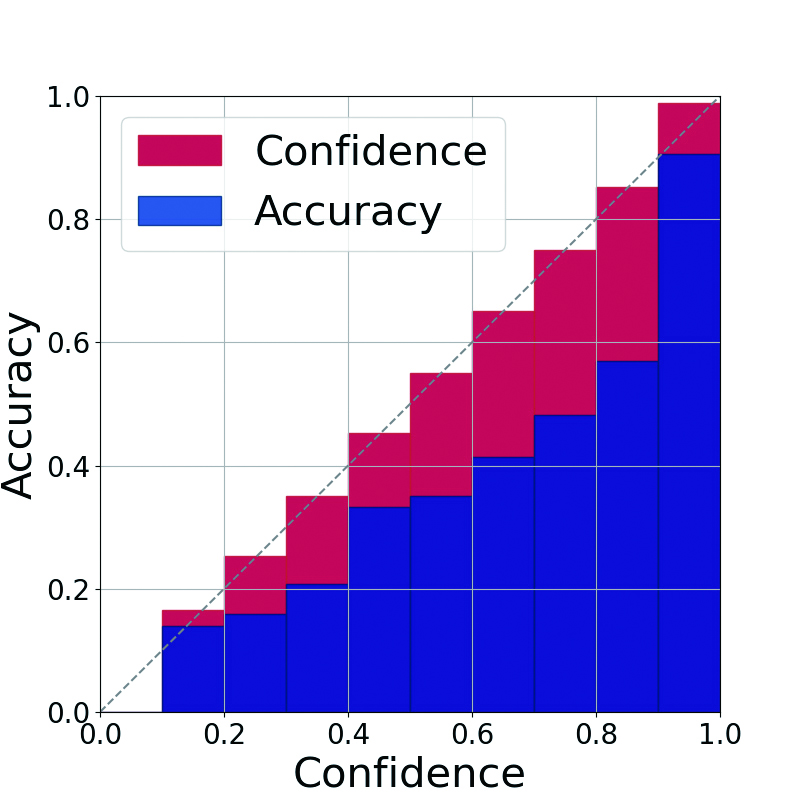}}
     \subfigure[Ours]{\includegraphics[width=80pt]{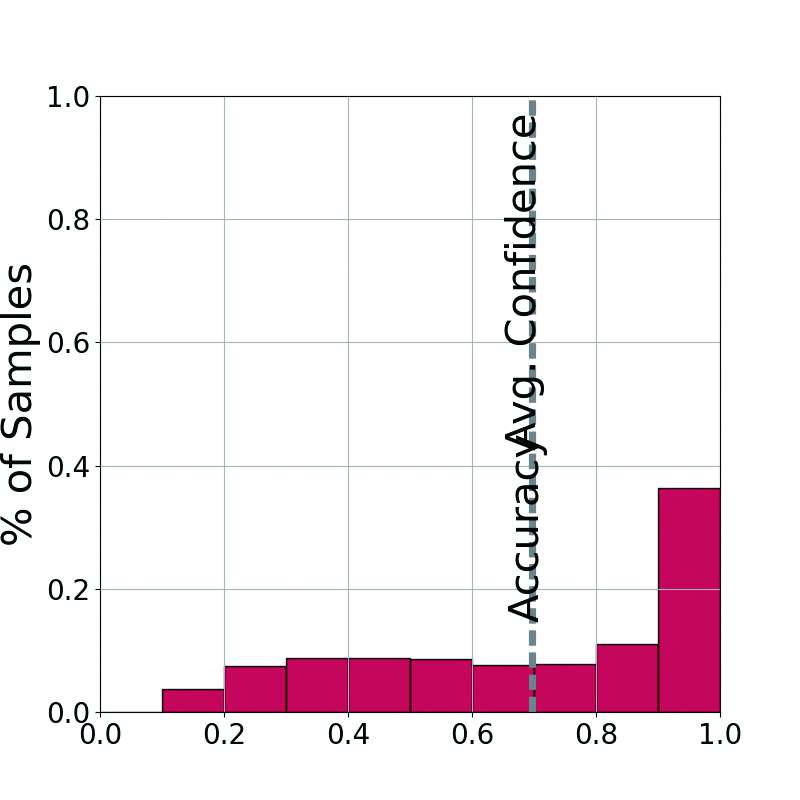}\includegraphics[width=80pt]{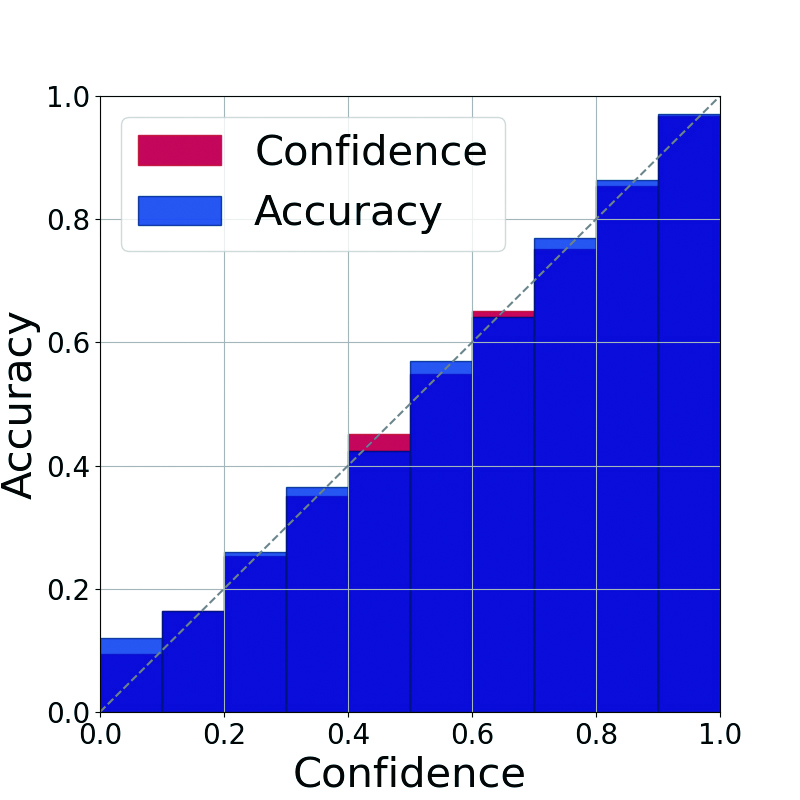}}
     \subfigure[Temp. Scaling]{\includegraphics[width=80pt]{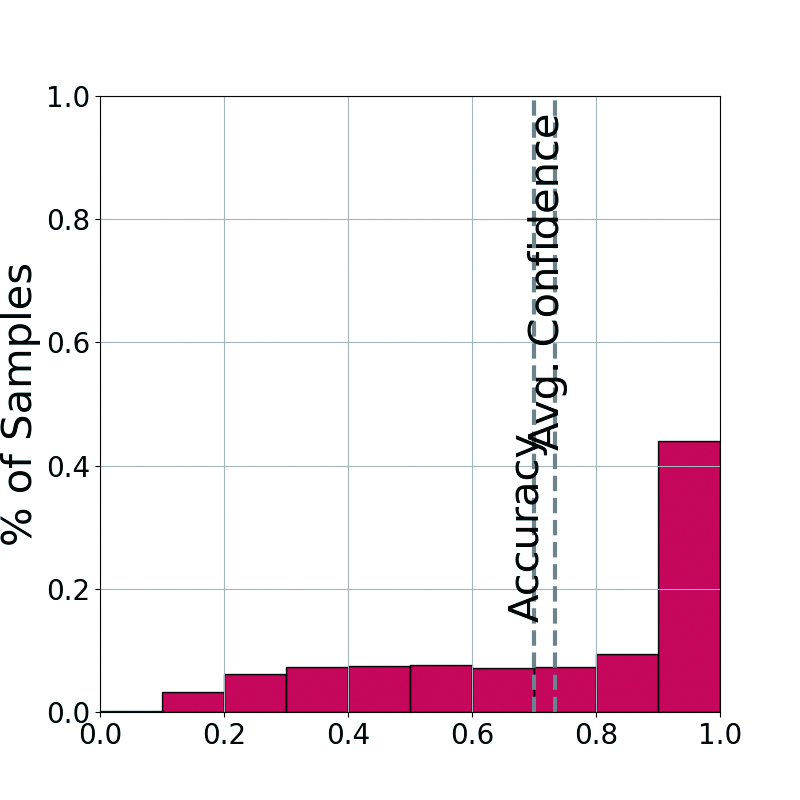}\includegraphics[width=80pt]{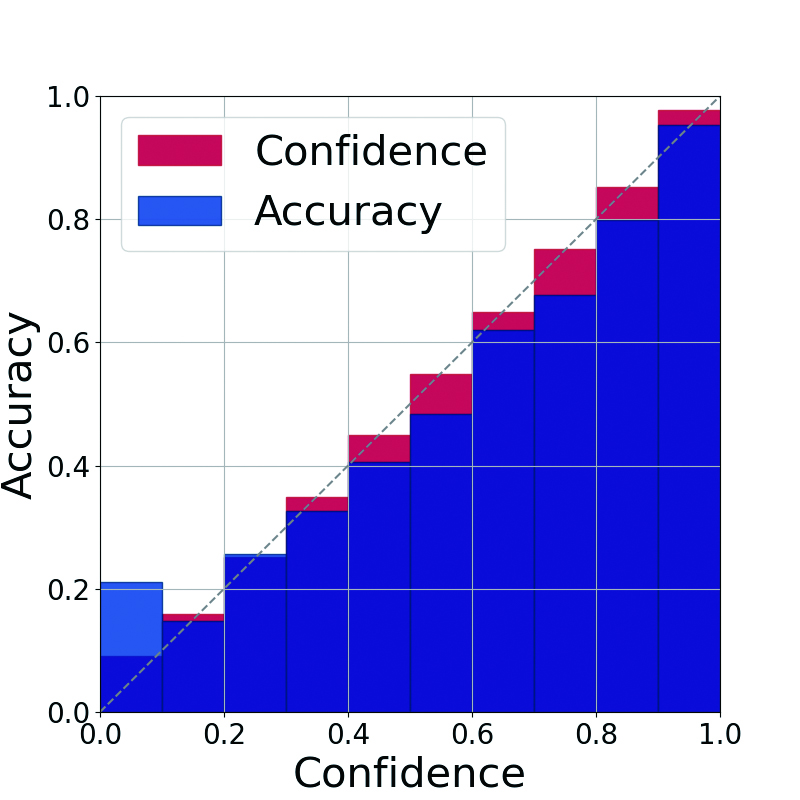}}
     \subfigure[Vector Scaling]{\includegraphics[width=80pt]{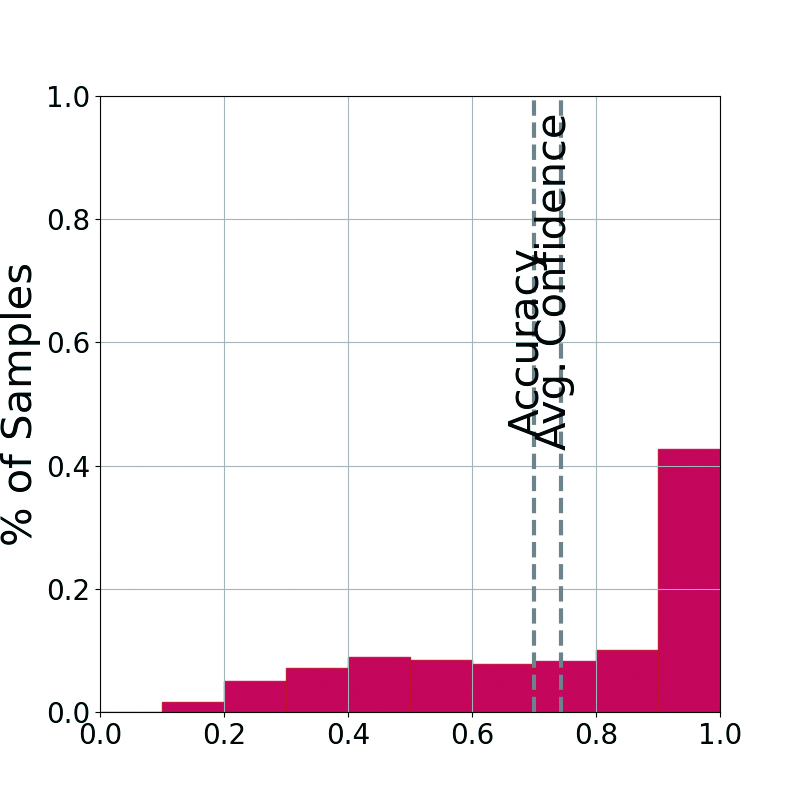}\includegraphics[width=80pt]{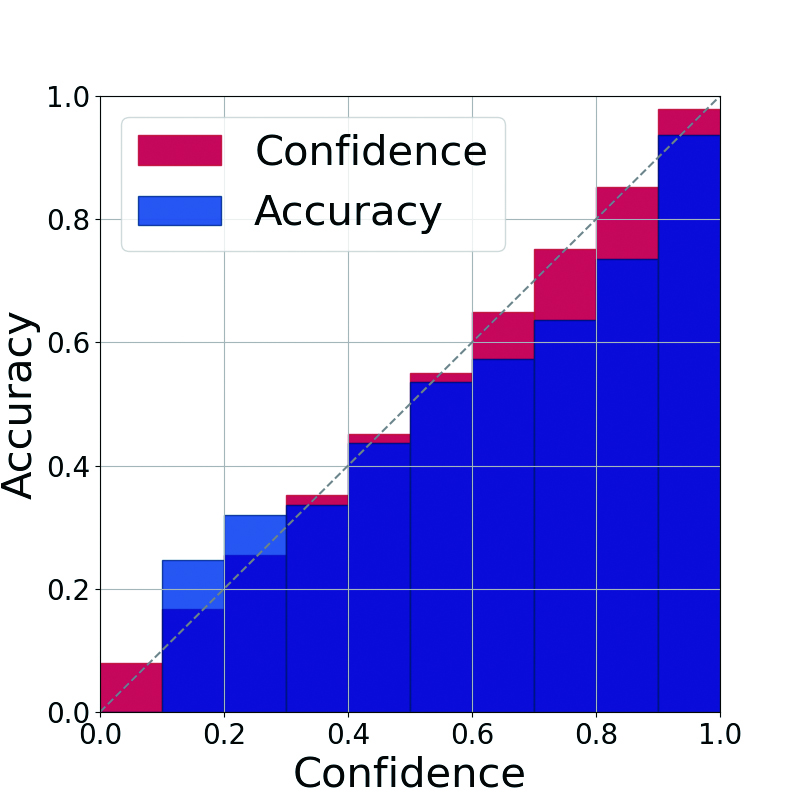}}
    \subfigure[TS-AvUC]{\includegraphics[width=80pt]{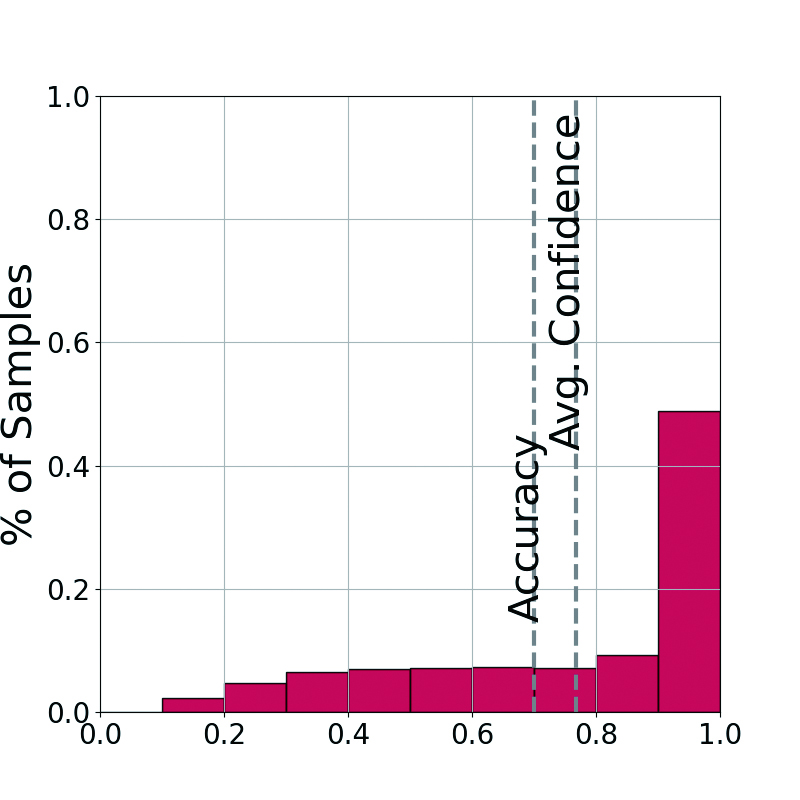}\includegraphics[width=80pt]{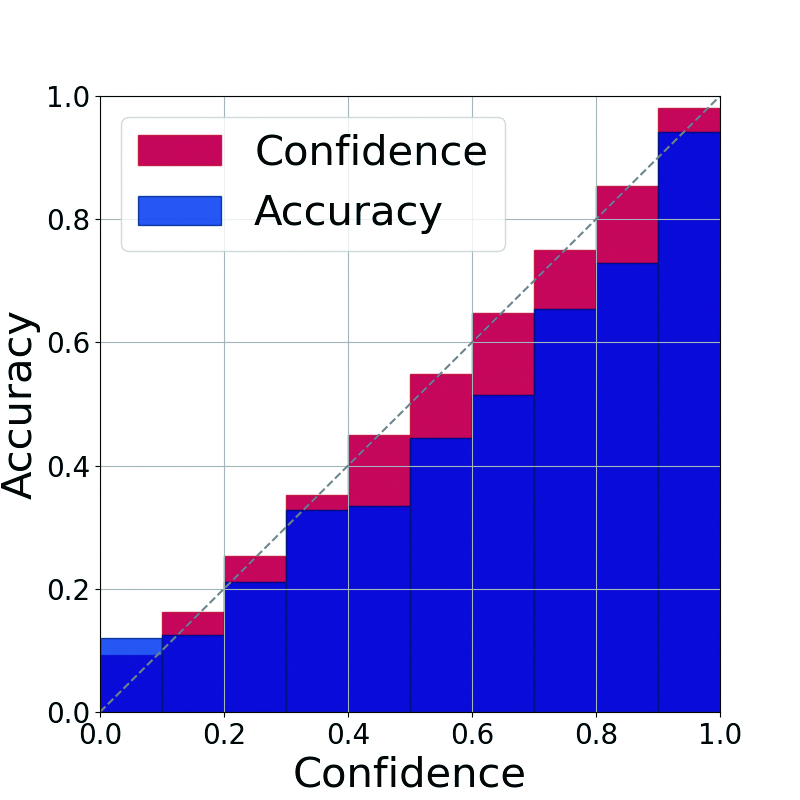}}
     \subfigure[Hist. Bin.]{\includegraphics[width=80pt]{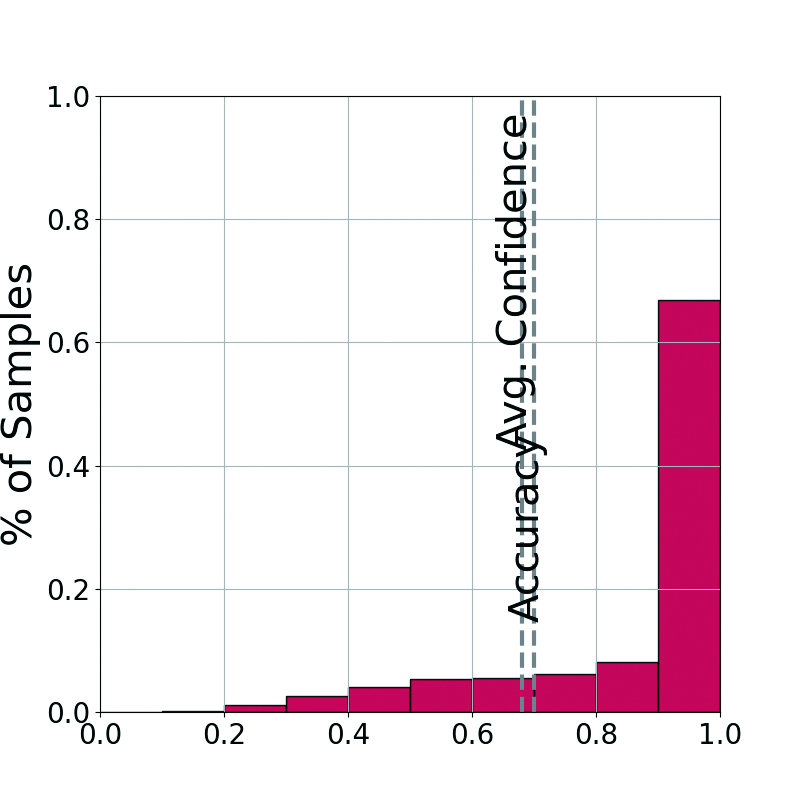}\includegraphics[width=80pt]{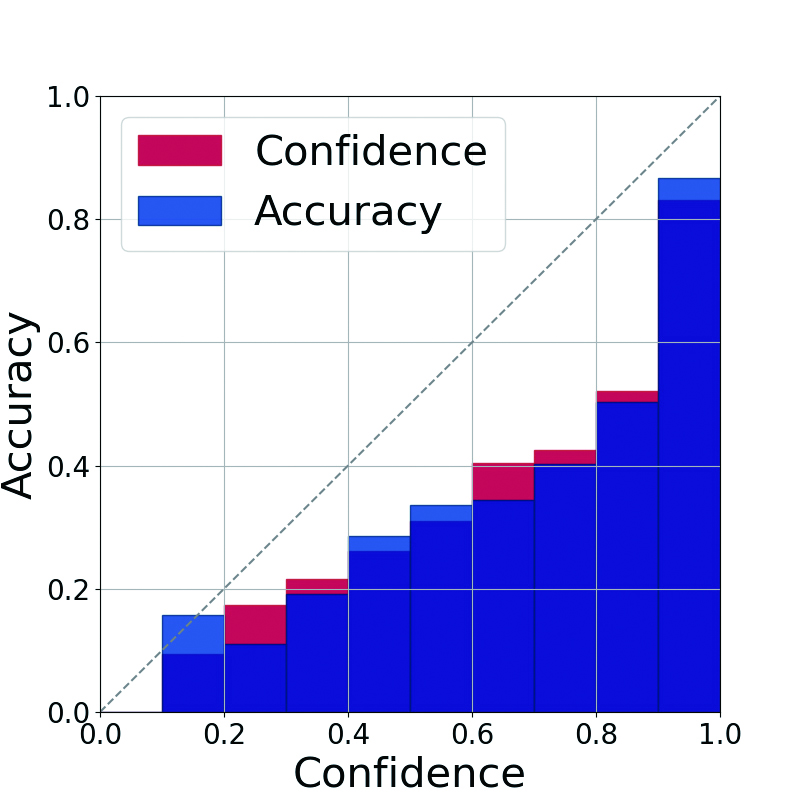}}
    \caption{\textbf{Confidence histograms and reliability diagrams for different post-hoc calibration methods with ResNet35 on CIFAR-100.} Confidence histograms display the sample count within each bin, whereas reliability diagrams illustrate the difference between the average confidence  (marked in red) and the accuracy (indicated in blue) in each bin.}
    \label{Confidence histograms and reliability diagrams for different calibration methods with ResNet 35 on CIFAR-100.}\label{Confidencedifferentmethods}
\end{figure*}

\subsection{Parameter optimization}
In this subsection, we address the challenge of optimizing the calibrator parameters from the desired calibration error. From a logical perspective, optimizing the calibrator with NLL does not differ from the optimization goal of the original classifier, Softmax cross entropy. However, it is essential to note that achieving high accuracy in classifiers using cross entropy as the objective and calibrating the model confidence represent distinct objectives \cite{compromiseacc}. There remains a bias in minimizing NLL compared to minimizing calibration error.

A straightforward approach is to utilize the calibration error as a loss function for optimizing the parametric calibration mapping. With calibration error as loss function, we treat the entire batch as a bin $D$ in each iteration of the optimization, randomly sampling a bin from the validation set. This approach helps alleviate the issue where finite data fail to fully reflect overall uncertainty. Modified  SCE (Square Calibration Error) is shown as follows.
\begin{equation}\label{eqsce}
    {l_{SCE}} = {\left( {acc\left( D \right) - conf\left( D \right)} \right)^2}
\end{equation}
where  $conf\left( D \right) = \frac{1}{{\left| D \right|}}\sum\nolimits_{{z^i} \in D} {\kappa \log \sum\nolimits_j^m {{e^{{{{g_j}\left( {{z_i}} \right)} \mathord{\left/
 {\vphantom {{{g_j}\left( {{z_i}} \right)} \kappa }} \right.
 \kern-\nulldelimiterspace} \kappa }}}} }$, ${g_j} = \frac{{{e^{{r_j}}}}}{{\sum\nolimits_{j = 1}^m {{e^{{r_j}}}} }}$ and  ${r_j}\left( z \right) = \frac{{{z_j}}}{{\gamma^2 {{\left\| z \right\|}_\rho } + \beta^2 }}$. The function for max confidence in each class is smoothed by sum-exp-up, and the coefficient $\kappa $  is set relatively small, such as $10^{-4}$, since the probability outputs are less than 1. The small $\kappa $ ensures the confidence closely approximates the max predictive probability. Furthermore, to satisfy Proposition \ref{Accuracy preserving} in optimization, we replace $\beta$  and $\gamma$  in the original  $\rho $-Norm Scaling with ${\beta ^2}$  and ${\gamma ^2}$  to the constraint that the hyperparameter is greater than zero, respectively.

However, the model uncertainty typically represents a statistic of the dataset, lacking the ability to adequately characterize individual samples. When directly using the calibration error as the optimization objective $\sum\nolimits_{i \in {D}} {{{\hat p}^i}} = \sum\nolimits_{i \in D} {\rm I} \left( {y_c^i = \hat y_c^i} \right)$, a substantial amount of sample-level information is lost in accurately characterizing the output-probability mapping $p^i \gets z^i$, which impedes finding a solution with strong generalization ability. So, we introduce instance-level KL divergence for the probability distribution between pre-calibration ${s}\left( {z}\right)$ and post-calibration ${g}\left( {z}\right)$, serving as a regularization technique to maintain the probability distribution similarity between probability distributions before and after calibration. This addition term ensures that the calibrated distribution retains specific distributional characteristics of the original one. It can provide important instance-level information for optimization.
\begin{equation}\label{eqkl}
{l_{KL}}= \sum\limits_{{z^i} \in D} {\sum\limits_{j = 1}^m {{g_j}({z^i})\left( {\log {g_j}\left( {{z^i}} \right) - \log {s_j}\left( {{z^i}} \right)} \right)} } 
\end{equation}
The final multi-level objective is represented below.
\begin{equation}\label{klplussce}
    l = l_{SCE} +  \alpha{l_{KL}}
\end{equation}

Relying solely on the bin-level SCE often leads to significant deviations between the pre- and post-calibration probability distributions due to its statistical nature, particularly when calibration structure has strong expressive capacity, as shown in Fig. \ref{Condidence distribution}. To mitigate this, we introduce the instance-level KL divergence of the original output distribution ${s}\left( {z}\right)$ concerning the calibrated probability distribution ${g}\left( {z}\right)$ as a regularization. When KL divergence converges to 0, ${g}\left( {z}\right)$ equals  ${s}\left( {z}\right)$. Therefore, ${s}\left( {z}\right)$ replaces the one-hot label in NLL, similar to label smoothing.

In parameter optimization, the proposed algorithm employs a two-stage optimization strategy.  In the outer loop, the algorithm conducts a grid search within the intervals $\left\{ {1,1.25, \ldots ,3} \right\}$ to determine $\rho$, using ECE as the metric. Meanwhile, in the inner loop, parameters $\gamma$ and $\beta$ are optimized by the small batch gradient-based methods with the proposed loss (\ref{klplussce}).

\begin{table*}[h]
\centering
\caption{The calibration performance of different post-hoc calibration methods.}\label{The calibration performance of different calibration methods}
\setlength{\tabcolsep}{0.7mm}{
\begin{tabular}{@{}ccccccccc@{}}
\toprule
Dataset                        & Model                     & Metric & Uncalibrated & Hist. Binning & TS & Vector Scaling & TS-AvUC & Ours   \\ \midrule
\multirow{3}{*}{CIFAR-100}     & \multirow{3}{*}{ResNet18} & ECE    & 0.160\scriptsize$\pm 0.025$       &0.025\scriptsize$\pm 0.006$       &0.033\scriptsize$\pm 0.006$      & 0.061\scriptsize$\pm 0.012$         & 0.028\scriptsize$\pm 0.004$         & \textbf{0.009}\footnotesize\textbf{(\textit{↓ 0.016}})\\
                               &                           & MCE    & 0.344\scriptsize$\pm 0.055$       &0.078\scriptsize$\pm 0.012$       & 0.059 \scriptsize $\pm 0.011$                    & 0.138 \scriptsize$\pm 0.022$        & \textbf{0.052}\footnotesize\textbf{(\textit{↓ 0.007}})         & 0.098$\pm 0.021$ \\
                               &                           & AdaECE & 0.160\scriptsize$\pm 0.023$       & -                                    & 0.030\scriptsize$\pm 0.007$        & 0.061\scriptsize$\pm 0.011$         & 0.027\scriptsize$\pm 0.006$         & \textbf{0.007}\footnotesize\textbf{(\textit{↓ 0.020}}) \\ \midrule
\multirow{3}{*}{CIFAR-100}     & \multirow{3}{*}{ResNet50} & ECE    & 0.186\scriptsize$\pm 0.031$       &0.025\scriptsize$\pm 0.004$       & 0.030\scriptsize$\pm 0.013$     & 0.073\scriptsize$\pm 0.021$        & 0.052\scriptsize$\pm 0.012$        & \textbf{0.007}\footnotesize\textbf{(\textit{↓ 0.018}}) \\
                               &                           & MCE    & 0.407\scriptsize$\pm 0.101$       &0.110\scriptsize$\pm 0.015$       & \textbf{0.091}\footnotesize\textbf{(\textit{↓ 0.009}})                  & 0.153\scriptsize$\pm 0.036$        & 0.116\scriptsize$\pm 0.021$        & 0.100$\pm 0.023$ \\
                               &                           & AdaECE & 0.186\scriptsize$\pm 0.029$       & -                                & 0.029\scriptsize$\pm 0.012$     & 0.071\scriptsize$\pm 0.028$        & 0.052\scriptsize$\pm 0.010$      & \textbf{0.006}\footnotesize\textbf{(\textit{↓ 0.023}}) \\ \midrule
\multirow{3}{*}{CIFAR-100}     & \multirow{3}{*}{VGG16}    & ECE    & 0.240\scriptsize$\pm 0.106$      &0.035\scriptsize$\pm 0.002$        & 0.029\scriptsize$\pm 0.003$        & 0.035\scriptsize$\pm 0.006$       & 0.044\scriptsize$\pm 0.008$       & \textbf{0.019}\footnotesize\textbf{(\textit{↓ 0.010}}) \\
                               &                           & MCE    & 0.508\scriptsize$\pm 0.151$      & \textbf{0.042}\footnotesize\textbf{(\textit{↓ 0.001}})                   & 0.093\scriptsize$\pm 0.029$        & 0.084\scriptsize$\pm 0.009$       & 0.101\scriptsize$\pm 0.026$       & 0.043$\pm 0.003$ \\
                               &                           & AdaECE & 0.240\scriptsize$\pm 0.106$      & -                                 & 0.029\scriptsize$\pm 0.004$        & 0.035\scriptsize$\pm 0.006$       & 0.044\scriptsize$\pm 0.008$        &\textbf{0.019}\footnotesize\textbf{(\textit{↓ 0.010}}) \\ \midrule
\multirow{3}{*}{CIFAR-10}      & \multirow{3}{*}{ResNet35} & ECE    & 0.054\scriptsize$\pm 0.010$      &0.011\scriptsize$\pm 0.001$      & 0.015\scriptsize$\pm 0.002$     & 0.014\scriptsize$\pm 0.003$       & 0.015\scriptsize$\pm 0.006$       & \textbf{0.007}\footnotesize\textbf{(\textit{↓ 0.004}}) \\
                               &                           & MCE    & 0.300\scriptsize$\pm 0.085$      &0.255\scriptsize$\pm 0.102$      & 0.121\scriptsize$\pm 0.026$     & \textbf{0.077}\footnotesize\textbf{(\textit{↓ 0.030}})                   & 0.121\scriptsize$\pm 0.021$       & 0.107$\pm 0.019$ \\
                               &                           & AdaECE & 0.054\scriptsize$\pm 0.011$      & -                               & 0.014\scriptsize$\pm 0.004$     & 0.013\scriptsize$\pm 0.002$       & 0.013\scriptsize$\pm 0.005$       &\textbf{0.010}\footnotesize\textbf{(\textit{↓ 0.003}}) \\ \midrule
\multirow{3}{*}{SVHN}      & \multirow{3}{*}{ResNet18}     & ECE    & 0.021\scriptsize$\pm 0.006$       &0.016\scriptsize$\pm 0.002$      & 0.009\scriptsize$\pm 0.003$       & \textbf{0.007}\footnotesize\textbf{(\textit{↓ 0.001}})                    & 0.010\scriptsize$\pm 0.003$       & 0.008\scriptsize\textbf{↑}$\pm 0.002$ \\
                               &                           & MCE    & 0.286\scriptsize$\pm 0.053$       & \textbf{0.251}\footnotesize\textbf{(\textit{↓ 0.035}})                & 0.313\scriptsize$\pm 0.052$       & 0.313\scriptsize$\pm 0.069$        & 0.315\scriptsize$\pm 0.080$       & 0.438$\pm 0.103$ \\
                               &                           & AdaECE & 0.021\scriptsize$\pm 0.006$       & -                               & 0.010\scriptsize$\pm 0.003$       & 0.009\scriptsize$\pm 0.002$                  & 0.013\scriptsize$\pm 0.005$       & \textbf{0.008}\footnotesize\textbf{(\textit{↓ 0.001}}) \\ \midrule
                               
\multirow{3}{*}{102 Flower} & \multirow{3}{*}{ResNet50} & ECE    & 0.101\scriptsize$\pm 0.018$       &0.084\scriptsize$\pm 0.012$      & 0.086\scriptsize$\pm 0.011$       &  0.093\scriptsize$\pm 0.015$       &  0.075\scriptsize$\pm 0.009$      & \textbf{0.046}\footnotesize\textbf{(\textit{↓ 0.029}}) \\
                               &                           & MCE    & 0.231\scriptsize$\pm 0.048$       &0.365\scriptsize$\pm 0.066$      & 0.180\scriptsize$\pm 0.041$       &  0.163\scriptsize$\pm 0.043$       &  0.165\scriptsize$\pm 0.044$      & \textbf{0.152}\footnotesize\textbf{(\textit{↓ 0.011}}) \\
                               &                           & AdaECE & 0.100\scriptsize$\pm 0.017$      & -                                & 0.089\scriptsize$\pm 0.012$       &  0.098\scriptsize$\pm 0.019$       &  0.079\scriptsize$\pm 0.011$      & \textbf{0.048}\footnotesize\textbf{(\textit{↓ 0.031}})  \\ \bottomrule
                               
\multirow{3}{*}{Tiny-ImageNet} & \multirow{3}{*}{ResNet35} & ECE    & 0.144\scriptsize$\pm 0.022$       &0.033\scriptsize$\pm 0.005$      & 0.017\scriptsize$\pm 0.003$       &  0.053\scriptsize$\pm 0.007$       &  0.017\scriptsize$\pm 0.003$      & \textbf{0.007}\footnotesize\textbf{(\textit{↓ 0.010}}) \\
                               &                           & MCE    & 0.236\scriptsize$\pm 0.052$       &0.055\scriptsize$\pm 0.016$      & 0.035\scriptsize$\pm 0.010$       &  0.093\scriptsize$\pm 0.021$       &  \textbf{0.030}\footnotesize\textbf{(\textit{↓ 0.001}})     & 0.031$\pm 0.004$ \\
                               &                           & AdaECE & 0.143\scriptsize$\pm 0.021$      & -                                & 0.017\scriptsize$\pm 0.004$       &  0.054\scriptsize$\pm 0.008$       &  0.016\scriptsize$\pm 0.002$      & \textbf{0.006}\footnotesize\textbf{(\textit{↓ 0.010}})  \\ \bottomrule
\end{tabular}}
\end{table*}

\begin{table*}
\centering
\caption{The ablation study of calibrator structure and optimization objective on CIFAR-100.}\label{The ablation study of calibrator structure on CIFAR-100}
\begin{tabular}{@{}cccccc@{}}
\toprule
\multirow{2}{*}{Model}    & \multirow{2}{*}{Metrics} & \multicolumn{2}{c}{NLL}        & \multicolumn{2}{c}{Ours}       \\ \cmidrule(l){3-6} 
                          &                          & Temp. Scaling & $\rho$-Norm Scaling & Temp. Scaling & $\rho$-Norm Scaling \\ \midrule
\multirow{2}{*}{ResNet35} & ECE                      & 0.026         & 0.011\footnotesize\textbf{(\textit{↓ 0.015}})          & 0.024         & 0.009\footnotesize\textbf{(\textit{↓ 0.015}})          \\
                          & AdaECE                   & 0.027         & 0.011\footnotesize\textbf{(\textit{↓ 0.016}})          & 0.024         & 0.007\footnotesize\textbf{(\textit{↓ 0.017}})          \\ \midrule
\multirow{2}{*}{ResNet50} & ECE                      & 0.048         & 0.006\footnotesize\textbf{(\textit{↓ 0.042}})          & 0.042         & 0.007\footnotesize\textbf{(\textit{↓ 0.035}})          \\
                          & AdaECE                   & 0.048         & 0.008\footnotesize\textbf{(\textit{↓ 0.040}})          & 0.042         & 0.006\footnotesize\textbf{(\textit{↓ 0.036}})          \\ \bottomrule
\end{tabular}
\end{table*}

\section{Experiments}\label{sec4}
We evaluate our methods on multiple DNNs, including ResNet and VGG series. Our experiments are conducted on SVHN, CIFAR-10/100, 102 Flower, and Tiny-ImageNet for post-hoc calibration performance. Different ablation experiments are designed to evaluate efficiency of the $\rho$-Norm Scaling calibration structure and the multi-level objective. In tables, the best results and relative improvements over $2^{nd}$ best result in each section are in bold. Results are averaged over five runs with different seeds.

\noindent\textbf{Baselines:} In experiments, we compare our methods with different calibration methods, such as non-parametric Hist. Binning, TS, Vector Scaling  \cite{matrixscaling}. Above all parametric structure are optimized by gradient descent based on NLL. TS-AvUC represents Tem. Scaling with NLL-AvUC as objective \cite{AVUO}. The  $\rho$ in  $\rho$-Norm Scaling is selected by grid search on  $\left\{ {1,1.25, \ldots ,3} \right\}$  and other parameter are optimized based on Eq.(\ref{klplussce}). In ablation experiments, different structures are optimized by different optimization objectives, such as NLL, SCE, AvUC and SB-ECE \cite{Softcalibration}. In Hist. Binning, ECE, MCE and AdaECE \cite{AdaECE}, the number of bins is 10. In all experiments for CIFAR-10/100, the learning rate was set to 0.1, the momentum to 0.9, the weight clipping to Norm=3, and the batch size to 128. The learning rate decreased to 10\% at 40\% and 80\% of the iterations. The weight decay was set to $10^{-4}$ and the iteration number was 200. For the Tiny-ImageNet, the learning rate was set to 0.01 and batch size was 64. The hyperparameter $\alpha$ is set to 1.\\

\begin{figure}[h]
    \centering
    \subfigure[Uncalibrated]{\includegraphics[width=75pt]{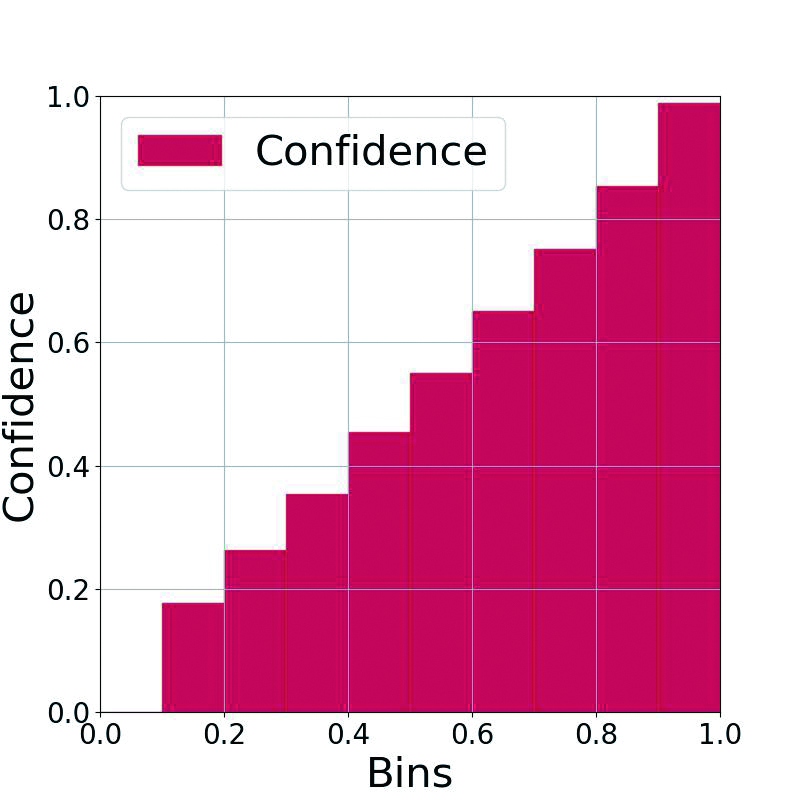}}
    \subfigure[SCE]{\includegraphics[width=75pt]{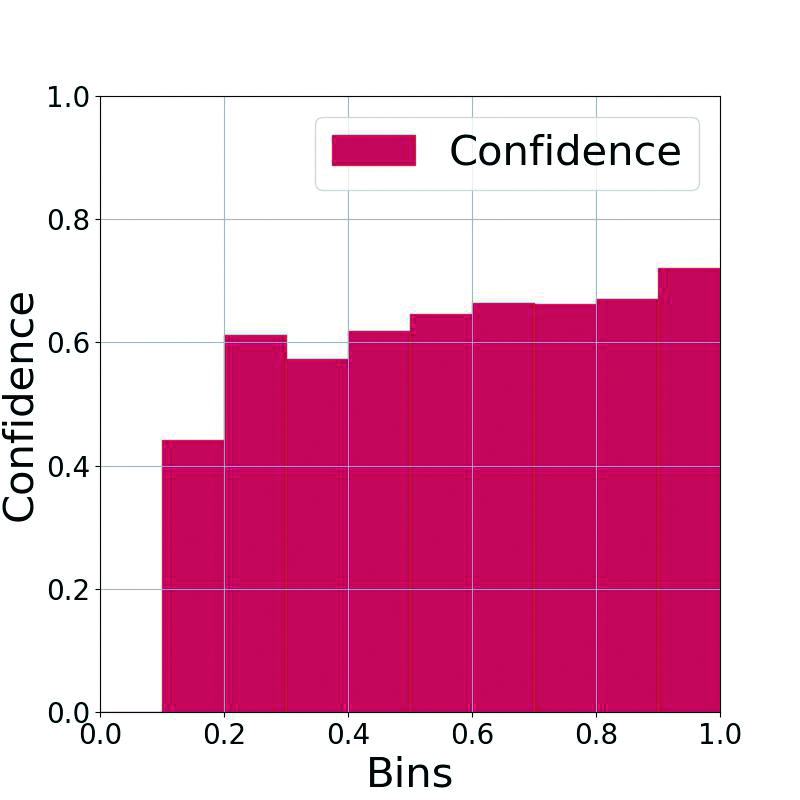}}
   \subfigure[SCE+KL]{\includegraphics[width=75pt]{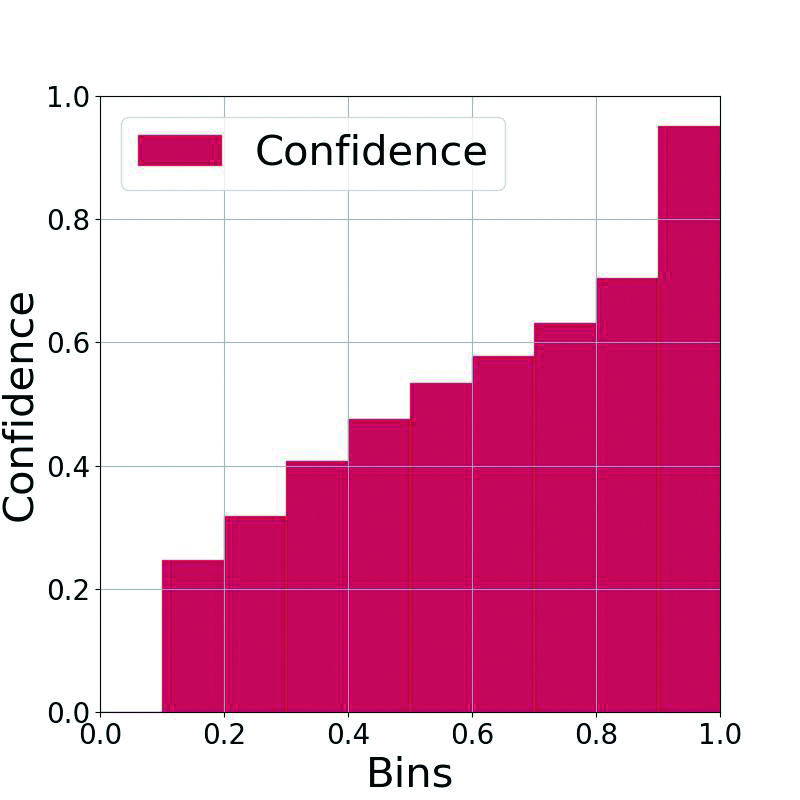}}
    \caption{ \textbf{Coincidence distribution of different optimization objective in Vector Scaling.} In (a), samples are categorized into bins based on confidence levels through Softmax. Each sample in (b) and (c) belongs to the same bin as in (a). Using sample-level SCE alone in post-calibration results in a significant deviation from the original distribution. This challenge is mitigated by the bin-level KL.}
    \label{Condidence distribution}
\end{figure}

\begin{table*}[t]
\centering
\caption{The ablation study with Vector Scaling on CIFAR-100.}\label{The ablation study with Vector Scaling on CIFAR-100}
\setlength{\tabcolsep}{1.5mm}{
\begin{tabular}{@{}cccccccccc@{}}
\toprule
Metrics & SCE   & KL    & SCE+KL & NLL   & NLL+KL & NLL-AvUC &SB-ECE & SB-ECE+KL & Uncalibrated \\ \midrule
ECE     & 0.173 & 0.161 & 0.041\footnotesize\textbf{(\textit{↓ 0.120}})  & 0.056 & 0.039\footnotesize\textbf{(\textit{↓ 0.008}}) & 0.047 & 0.156 & 0.039 \footnotesize\textbf{(\textit{↓ 0.117}}) & 0.172        \\
AdaECE  & 0.172 & 0.161 & 0.043\footnotesize\textbf{(\textit{↓ 0.118}})  & 0.053 & 0.040\footnotesize\textbf{(\textit{↓ 0.005}}) &0.045 & 0.157 & 0.036 \footnotesize\textbf{(\textit{↓ 0.121}}) & 0.172        \\
KL      & 0.109 & 0.001 & 0.003  & 0.006 & 0.004 & 0.002 & 0.097 & 0.002 & -            \\ \bottomrule
\end{tabular}}
\end{table*}

\begin{table*}[h]
\centering
\caption{The calibration performance on ResNet35 of different norm in $\rho$-Norm Scaling.}\label{The calibration performance on ResNet35 of different norm in  -norm scaling}
\setlength{\tabcolsep}{1.5mm}{
\begin{tabular}{@{}ccccccccc@{}}
\toprule
\multirow{2}{*}{Dataset}       & \multirow{2}{*}{Metrics} & \multicolumn{5}{c}{ Different $\rho$ in $\rho$-Norm Scaling}                 & \multirow{2}{*}{Uncalibrated} & \multirow{2}{*}{Temp. Scaling} \\ \cmidrule(lr){3-7}
                               &                          & 1.5   & 1.75  & 2     & 2.25  & 2.5   &                               &                                \\ \midrule
\multirow{2}{*}{CIFAR-100}     & ECE                      & 0.028 & \textbf{0.006}\footnotesize\textbf{(\textit{↓ 0.003}})& 0.010 & 0.009 & 0.010 & 0.172                         & 0.026                          \\
                               & AdaECE                   & 0.028 & 0.008 & 0.009 & \textbf{0.007}\footnotesize\textbf{(\textit{↓ 0.001}}) & 0.009 & 0.172                         & 0.027                          \\ \midrule
\multirow{2}{*}{Tiny-ImageNet} & ECE                      & 0.052 & \textbf{0.007}\footnotesize\textbf{(\textit{↓ 0.011}}) & 0.018 & 0.040 & 0.044 & 0.144                         & 0.017                          \\
                               & AdaECE                   & 0.051 & \textbf{0.006}\footnotesize\textbf{(\textit{↓ 0.012}}) & 0.018 & 0.041 & 0.044 & 0.143                         & 0.017                          \\ \bottomrule
\end{tabular}}
\end{table*}

\noindent\textbf{The efficiency of our method:}
Table \ref{The calibration performance of different calibration methods} presents the outcomes of various calibration techniques. Our method substantially enhances the performance of post-calibration in both ECE and AdaECE, outperforming classical methods. However, $\rho$-Norm Scaling does not yield superior results in MCE. The confidence histograms and reliability diagrams depicted in Fig. \ref{Confidencedifferentmethods} reveal a bias in bin $\left[ {0,0.1} \right]$  with a minimal number of samples, which does not significantly impact the overall calibration outcomes in ECE and AdaECE. TS, utilizing a single hyperparameter to control the smoothing of uncertainty distribution, achieves superior results compared to post-processing calibrations employing multiple hyperparameters like Vector Scaling. Furthermore, both TS and our proposed method demonstrate accuracy-preserving property, preserving the order of probability outputs across different categories for individual samples \cite{intraorder}. The notably smaller calibration errors of these mapping, in contrast to Vector Scaling, underscore the pivotal role of maintaining decision invariance as a fundamental prerequisite in calibrator design.\\

\noindent\textbf{The ablation study of calibrator structure:}
To mitigate the influence of the loss function on the experimental outcomes, we conducted ablation experiments and presented the results in Table \ref{The ablation study of calibrator structure on CIFAR-100}. Notably, in Table \ref{The ablation study of calibrator structure on CIFAR-100},  $\rho$-Norm Scaling continues to enhance calibration performance when compared to TS under the same optimization objective. This observation suggests that the supervised optimization using Softmax cross entropy as the objective leads to a larger amplitude in model output, negatively impacting calibration. Furthermore, it fails to ensure both classifier accuracy and uncertainty estimates derived from Softmax mapping. Consequently, $\rho$-Norm Scaling can realize better calibration performance than TS. \\

\noindent\textbf{The ablation study of optimization objective:}
The KL divergence regularity introduces instance-level information into calibrator optimization, ensuring that the calibrated probability retains certain properties of the uncalibrated distribution, as emphasized by the results in Table \ref{The ablation study with Vector Scaling on CIFAR-100}. The bin-level calibration error as a statistical representation of collective binning properties, compressing sample-level information significantly. Furthermore, Vector Scaling offers a broader assumption space and greater expressive power in comparison to Temp. Scaling. When SCE or SB-ECE is used solely as the loss function, the calibrated results deviate markedly from the original results as shown in Fig. \ref{Condidence distribution} and the KL divergence remains relatively large. In addition, using KL divergence alone does not yield improved results. However, better outcomes are achieved when bin-level loss and instance-level loss jointly optimize parameters. KL divergence, acting as a regularity term, guides the calibrated model to learn distributions mirroring the properties of the original distributions, while SCE fine-tunes the mapping parameters and refines the calibrator model. \\

\noindent\textbf{The ablation study of different norms:}
Table \ref{The calibration performance on ResNet35 of different norm in  -norm scaling} displays calibration results across various norms. The data illustrates that smoother outcomes are achieved when $\rho$ is close to 2, though $\rho$ does not guarantee optimality. Different learning paradigms enable the exploration of diverse spaces, facilitating the acquisition of more precise output-probability mappings. This reflects the significance of searching for the appropriate norm, rather than directly replacing it with the conventional Euclidean norm.

\section{Conclusion}
The desired calibration metric, based on sample set statistics, captures the dataset general characteristics but overlooks sample-level nuances. Relying solely on the calibration error as the optimization objective fails to yield a well-generalized output-probability mapping within a broad assumption space. Consequently, integrating specific priori knowledge becomes imperative when designing and optimizing the post-hoc calibrator. In this paper, we introduce a $\rho$-Norm Scaling to mitigate the adverse impact of amplified output amplitude in supervised learning while preserving accuracy. Simultaneously, an instance-level probability distribution regularization is proposed in the optimization, which incorporates specific priori knowledge and emphasizes the need for the uncertainty distribution after calibration to keep some characteristics of the pre-calibration distribution. The experimental results show the significant enhancement in uncertainty calibration performance through $\rho$-Norm Scaling and multi-level objective. They also underscore the necessity for precise calibrator design to guide the model effectively in learning an ideal calibration mapping. 
\bibliography{aaai25}

\appendix
\section*{Appendix}
\subsection{Proof of Proposition \ref{Bound of RMSNorm Softmax}}
To calculate the extreme value of $\frac{{{e^{{r_c}}}}}{{\sum\nolimits_{j = 1}^m {{e^{{r_j}}}} }}$  where  ${r_j} = \frac{{{z_j}}}{{\gamma {{\left( {\sum\nolimits_{j = 1}^m {z_j^\rho } } \right)}^{1/\rho }}}}$, we consider the extreme value of the following function with restrictions.
\begin{equation}
    p = \frac{{{e^{{r_c}}}}}{{\sum\nolimits_{j = 1}^m {{e^{{r_j}}}} }} \\ \begin{array}{*{20}{c}}
{s.t.}&{\sum\nolimits_{j = 1}^m {r_j^\rho  = \frac{1}{{{\gamma ^\rho }}}} }
\end{array}
\end{equation}
We construct the Lagrangian function as follows.
\begin{equation}
    L = \ln \frac{{{e^{{r_c}}}}}{{\sum\nolimits_{j = 1}^m {{e^{{r_j}}}} }} + \lambda (\sum\nolimits_{j = 1}^m {r_j^\rho  - \frac{1}{{{\gamma ^\rho }}}} )
\end{equation}
Let the derivative of this function be 0.
\begin{equation}\label{eq10}
    \frac{{\partial L}}{{\partial \lambda }} = \sum\nolimits_{j = 1}^m {r_j^\rho  -  - \frac{1}{{{\gamma ^\rho }}}}  = 0
\end{equation}
\begin{equation}\label{eq11}
    \frac{{\partial L}}{{{r_c}}} = 1 - \frac{{{e^{{r_c}}}}}{{\sum\nolimits_{k = 1}^m {{e^{{r_k}}}} }} \pm \rho \lambda {r_c}^{\rho  - 1} = 0
\end{equation}
\begin{equation}\label{eq12}
    \frac{{\partial L}}{{{r_j}}} = \frac{{{e^{{r_j}}}}}{{\sum\nolimits_{k = 1}^m {{e^{{r_k}}}} }} \mp \rho \lambda {r_j}^{\rho  - 1} = 0,j \ne c
\end{equation}
From (\ref{eq12}), we can obtain
\begin{equation}
    \lambda  = \frac{{{e^{{r_j}}}}}{{\rho {r_j}^{\rho  - 1}\sum\nolimits_{k = 1}^m {{e^{{r_k}}}} }},j \ne c
\end{equation}
${r_j},j = 1, \ldots ,m,j \ne c$  have same sign when function obtain the extreme point. We can obtain 
\begin{equation}\label{eq14}
    {r_1} = {r_2} = {r_j},j = 1, \ldots ,m,j \ne c
\end{equation}
Adding (\ref{eq11}) and (\ref{eq12}) together, we can get
\begin{equation}\label{eq15}
    {r_1}^{\rho  - 1} + {r_2}^{\rho  - 1} +  \cdots  + {r_m}^{\rho  - 1} = {r_c}^{\rho  - 1}
\end{equation}
Combining (\ref{eq10}), (\ref{eq14}) and (\ref{eq15}), we can obtain extreme value point

\begin{equation}
    \left\{ {\begin{array}{*{20}{c}}
{{r_c} = \frac{{{{\left( {m - 1} \right)}^{\frac{\rho }{{\rho  - 1}}}}}}{{\gamma {{\left( {\left( {m - 1} \right) + {{\left( {m - 1} \right)}^{\frac{\rho }{{\rho  - 1}}}}} \right)}^{\frac{1}{\rho }}}}}}\\
{{r_j} =  - \frac{1}{{\gamma {{\left( {\left( {m - 1} \right) + {{\left( {m - 1} \right)}^{\frac{\rho }{{\rho  - 1}}}}} \right)}^{\frac{1}{\rho }}}}}j \ne c}
\end{array}} \right. 
 \mbox{or} \end{equation}
\begin{equation}
\left\{ {\begin{array}{*{20}{c}}
{{r_c} =  - \frac{{{{\left( {m - 1} \right)}^{\frac{\rho }{{\rho  - 1}}}}}}{{\gamma {{\left( {\left( {m - 1} \right) + {{\left( {m - 1} \right)}^{\frac{\rho }{{\rho  - 1}}}}} \right)}^{\frac{1}{\rho }}}}}}\\
{{r_j} = \frac{1}{{\gamma {{\left( {\left( {m - 1} \right) + {{\left( {m - 1} \right)}^{\frac{\rho }{{\rho  - 1}}}}} \right)}^{\frac{1}{\rho }}}}}j \ne c}
\end{array}} \right.
\end{equation}

\begin{table*}[t]
\centering\caption{Ablation experiments of $\alpha$ of multi-level objective on $\rho$-Norm Scaling calibration.}
\begin{tabular}{ccccccccc}
\hline
Dataset                    & Model                     & Metric & 0     & 0.01  & 0.1   & 1     & 10    & 100   \\ \hline
\multirow{2}{*}{CIFAR-100} & \multirow{2}{*}{ResNet35} & ECE    & 0.173 & 0.091 & 0.052 & 0.041 & 0.101 & 0.172 \\
                           &                           & KL     & 0.106 & 0.086 & 0.023 & 0.003 & 0.002 & 0.001 \\ \hline
\multirow{2}{*}{102 Flower} & \multirow{2}{*}{ResNet50} & ECE    & 0.100 & 0.083 & 0.059 & 0.048 & 0.053 & 0.102 \\
                           &                           & KL     & 0.123 & 0.072 & 0.031 & 0.002 & 0.002 & 0.001 \\ \hline
\end{tabular}\label{Tab5}
\end{table*}

\begin{equation}
\begin{array}{l}
\max p = \max \left\{ {\frac{1}{{\left( {m - 1} \right){e^{ - \frac{1}{\gamma }{{\left( {\frac{1}{{{{\left( {m - 1} \right)}^{{1 \mathord{\left/
 {\vphantom {1 {\rho  - 1}}} \right.
 \kern-\nulldelimiterspace} {\rho  - 1}}}}}} + 1} \right)}^{{{\rho  - 1} \mathord{\left/
 {\vphantom {{\rho  - 1} \rho }} \right.
 \kern-\nulldelimiterspace} \rho }}}}} + 1}},} \right.\\
\left. {\frac{{{e^{\frac{1}{\gamma }}}}}{{\left( {m - 1} \right) + {e^{\frac{1}{\gamma }}}}},\frac{1}{{\left( {m - 1} \right) + {e^{ - \frac{1}{\gamma }}}}}} \right\}
\end{array}
\end{equation}

$\frac{{{e^{\frac{1}{\gamma }}}}}{{\left( {m - 1} \right) + {e^{\frac{1}{\gamma }}}}} > \frac{1}{{\left( {m - 1} \right) + {e^{ - \frac{1}{\gamma }}}}}$ holds when $\gamma  > 0$.
\begin{equation}
\begin{array}{c}
\max p = \max \left\{ {\frac{1}{{\left( {m - 1} \right){e^{ - \frac{1}{\gamma }{{\left( {\frac{1}{{{{\left( {m - 1} \right)}^{{1 \mathord{\left/
 {\vphantom {1 {\rho  - 1}}} \right.
 \kern-\nulldelimiterspace} {\rho  - 1}}}}}} + 1} \right)}^{{{\rho  - 1} \mathord{\left/
 {\vphantom {{\rho  - 1} \rho }} \right.
 \kern-\nulldelimiterspace} \rho }}}}} + 1}},\frac{1}{{\left( {m - 1} \right){e^{ - \frac{1}{\gamma }}} + 1}}} \right\}\\
 = \frac{1}{{\left( {m - 1} \right){e^{ - \frac{1}{\gamma }{{\left( {\frac{1}{{{{\left( {m - 1} \right)}^{{1 \mathord{\left/
 {\vphantom {1 {\rho  - 1}}} \right.
 \kern-\nulldelimiterspace} {\rho  - 1}}}}}} + 1} \right)}^{{{\rho  - 1} \mathord{\left/
 {\vphantom {{\rho  - 1} \rho }} \right.
 \kern-\nulldelimiterspace} \rho }}}}} + 1}}
\end{array}
\end{equation}
\begin{equation}
\begin{array}{l}
\min p = \min \left\{ {\frac{1}{{\left( {m - 1} \right){e^{ - \frac{1}{\gamma }{{\left( {\frac{1}{{{{\left( {m - 1} \right)}^{{1 \mathord{\left/
 {\vphantom {1 {\rho  - 1}}} \right.
 \kern-\nulldelimiterspace} {\rho  - 1}}}}}} + 1} \right)}^{{{\rho  - 1} \mathord{\left/
 {\vphantom {{\rho  - 1} \rho }} \right.
 \kern-\nulldelimiterspace} \rho }}}}} + 1}},} \right.\\
\left. {\frac{{{e^{ - \frac{1}{\gamma }}}}}{{\left( {m - 1} \right) + {e^{ - \frac{1}{\gamma }}}}},\frac{1}{{\left( {m - 1} \right) + {e^{\frac{1}{\gamma }}}}}} \right\}
\end{array}
\end{equation}
And $\frac{{{e^{ - \frac{1}{\gamma }}}}}{{\left( {m - 1} \right) + {e^{ - \frac{1}{\gamma }}}}} < \frac{1}{{\left( {m - 1} \right) + {e^{\frac{1}{\gamma }}}}}$ holds when  $\gamma  > 0$. 
\begin{equation}
    \begin{array}{c}
\min p = \min \left\{ {\frac{1}{{\left( {m - 1} \right){e^{ - \frac{1}{\gamma }{{\left( {\frac{1}{{{{\left( {m - 1} \right)}^{{1 \mathord{\left/
 {\vphantom {1 {\rho  - 1}}} \right.
 \kern-\nulldelimiterspace} {\rho  - 1}}}}}} + 1} \right)}^{{{\rho  - 1} \mathord{\left/
 {\vphantom {{\rho  - 1} \rho }} \right.
 \kern-\nulldelimiterspace} \rho }}}}} + 1}},\frac{1}{{\left( {m - 1} \right){e^{\frac{1}{\gamma }}} + 1}}} \right\}\\
 = \frac{1}{{\left( {m - 1} \right){e^{ - \frac{1}{\gamma }{{\left( {\frac{1}{{{{\left( {m - 1} \right)}^{{1 \mathord{\left/
 {\vphantom {1 {\rho  - 1}}} \right.
 \kern-\nulldelimiterspace} {\rho  - 1}}}}}} + 1} \right)}^{{{\rho  - 1} \mathord{\left/
 {\vphantom {{\rho  - 1} \rho }} \right.
 \kern-\nulldelimiterspace} \rho }}}}} + 1}}
\end{array}
\end{equation}
\begin{equation}
 \begin{split}
     \frac{1}{{\left( {m - 1} \right){e^{\frac{1}{\gamma }{{\left( {\frac{1}{{{{\left( {m - 1} \right)}^{{1 \mathord{\left/
 {\vphantom {1 {\rho  - 1}}} \right.
 \kern-\nulldelimiterspace} {\rho  - 1}}}}}} + 1} \right)}^{{{\rho  - 1} \mathord{\left/
 {\vphantom {{\rho  - 1} \rho }} \right.
 \kern-\nulldelimiterspace} \rho }}}}} + 1}} \le g \\ \le \frac{1}{{\left( {m - 1} \right){e^{ - \frac{1}{\gamma }{{\left( {\frac{1}{{{{\left( {m - 1} \right)}^{{1 \mathord{\left/
 {\vphantom {1 {\rho  - 1}}} \right.
 \kern-\nulldelimiterspace} {\rho  - 1}}}}}} + 1} \right)}^{{{\rho  - 1} \mathord{\left/
 {\vphantom {{\rho  - 1} \rho }} \right.
 \kern-\nulldelimiterspace} \rho }}}}} + 1}}
 \end{split}
\end{equation}

\subsection{Proof of Proposition \ref{Accuracy preserving}}
For any model output  ${z_1},{z_2}, \ldots ,{z_m}$, $\gamma  > 0$  and $\beta  > 0$ ,  $\sum\nolimits_{j = 1}^m {{e^{{r_j}}}}  > 0$ and $\gamma {\left\| {\rm{z}} \right\|_\rho } + \beta  > 0$  in the ${p_j} = \frac{{{e^{{r_j}}}}}{{\sum\nolimits_{j = 1}^m {{e^{{r_j}}}} }}$ Softmax with  $\rho $-Norm Scaling  ${r_j} = \frac{{{z_j}}}{{\gamma \sqrt {\sum\nolimits_{j = 1}^m {z_j^2} }  + \beta }}$ are same. $p\left( {{z_j}} \right) = {{{e^{{{{z_j}} \mathord{\left/
 {\vphantom {{{z_j}} a}} \right.
 \kern-\nulldelimiterspace} a}}}} \mathord{\left/
 {\vphantom {{{e^{{{{z_j}} \mathord{\left/
 {\vphantom {{{z_j}} a}} \right.
 \kern-\nulldelimiterspace} a}}}} b}} \right.
 \kern-\nulldelimiterspace} b}$  is monotonically increasing function. So,  $\rho $-Norm Scaling satisfies the strictly order-preserving property for different class output, such that ${z_j} > {z_q} \Rightarrow {p_j} > {p_q}$ , then the model accuracy  keeps unchanged.
\subsection{Proof of Proposition \ref{Accuracy preserving2}}
For any model output  ${z_1},{z_2}, \ldots ,{z_m}$, and function  $\sigma \left( z \right) > 0$, if  ${z_j} > {z_q}$, ${z_j}\sigma \left( {{z_1},{z_2},...,{z_m}} \right) > {z_q}\sigma \left( {{z_1},{z_2},...,{z_m}} \right)$ holds. So Softmax  ${p_j}\left( z \right) = \frac{{{e^{{z_j}\sigma \left( z \right)}}}}{{\sum\nolimits_{j = 1}^m {{e^{{z_j}\sigma \left( z \right)}}} }}$ satisfies the strictly intra order-preserving property for different class output, such that ${z_j} > {z_q} \Rightarrow {p_j} > {p_q}$, then preserves the accuracy.
\subsection{Experiments compute resources}
For experiments, we utilized compute resources featuring an NVIDIA A100 GPU with PCIe interface and 40GB memory capacity, accompanied by PyTorch version 1.7.0 with CUDA version 11.0. The computational backbone was supported by an Intel Xeon Gold 6278C processor.
\subsection{Limitations and future works} \label{Limitations and future works}

The proposed calibration structure, $\rho$-Norm Scaling, enhances the expressiveness of TS in the structural design. By adding the correlation term of magnitude, our approach leverages the positive relationship between output magnitude and confidence, thereby refining the model's expressiveness. Nonetheless, further investigation into a more precise representation grounded in decision invariance is warranted, as described in Proposition \ref{Accuracy preserving2}. Additionally, we employ the grid search method to optimize the parameter $\rho$, leading to bi-optimization and heightened optimization complexity. While this method facilitates swift optimization due to the few optimization parameters, it exhibits increased optimization complexity in contrast to TS.

We adopt a multi-level optimization objective in our optimization approach. A regular term leveraging the similarity between pre- and post-calibration distributions is integrated to address the lack of instance-level information inherent in bin-level expected calibration error optimization. However, an excessively large value for $\alpha$ may adversely affect calibration results, as depicted in Tab. \ref{Tab5}. Thus, determining an appropriate value for $\alpha$ remains a challenge. Additionally, the impact of this calibration structure on TS is notably inferior compared to calibration models with more parameters, such as Vector Scaling.The regular term has limited efficacy in optimization problems with few parameters, but it proves effective in well-calibrated models featuring a larger parameter space.

\section*{Reproducibility Checklist}
This paper:
    \begin{itemize}
        \item Includes a conceptual outline and/or pseudocode description of AI methods introduced (yes)
        \item Clearly delineates statements that are opinions, hypothesis, and speculation from objective facts and results (yes)
        \item Provides well marked pedagogical references for less-familiare readers to gain background necessary to replicate the paper (yes)
    \end{itemize}
Does this paper make theoretical contributions? (yes)

\noindent If yes, please complete the list below.
    \begin{itemize}
       \item All assumptions and restrictions are stated clearly and formally. (yes)
        \item All novel claims are stated formally (e.g., in theorem statements). (yes)
        \item Proofs of all novel claims are included. (yes)
        \item Proof sketches or intuitions are given for complex and/or novel results. (yes)
        \item Appropriate citations to theoretical tools used are given. (NA)
        \item All theoretical claims are demonstrated empirically to hold. (yes)
        \item All experimental code used to eliminate or disprove claims is included. (NA)
    \end{itemize}
Does this paper include computational experiments? (yes)
\noindent If yes, please complete the list below.
    \begin{itemize}
        \item Any code required for pre-processing data is included in the appendix. (yes).
        \item All source code required for conducting and analyzing the experiments is included in a code appendix. (no)
        \item All source code required for conducting and analyzing the experiments will be made publicly available upon publication of the paper with a license that allows free usage for research purposes. (yes)
        \item All source code implementing new methods have comments detailing the implementation, with references to the paper where each step comes from (yes)
        \item If an algorithm depends on randomness, then the method used for setting seeds is described in a way sufficient to allow replication of results. (yes)
        \item This paper specifies the computing infrastructure used for running experiments (hardware and software), including GPU/CPU models; amount of memory; operating system; names and versions of relevant software libraries and frameworks. (yes)
        \item This paper formally describes evaluation metrics used and explains the motivation for choosing these metrics. (partial)
        \item This paper states the number of algorithm runs used to compute each reported result. (yes)
        \item Analysis of experiments goes beyond single-dimensional summaries of performance (e.g., average; median) to include measures of variation, confidence, or other distributional information. (yes)
        \item The significance of any improvement or decrease in performance is judged using appropriate statistical tests (e.g., Wilcoxon signed-rank). (no)
        \item This paper lists all final (hyper-)parameters used for each model/algorithm in the paper’s experiments. (yes)
        \item This paper states the number and range of values tried per (hyper-) parameter during development of the paper, along with the criterion used for selecting the final parameter setting. (yes)
    \end{itemize}

\end{document}